\documentclass{amsart}
\usepackage{amsmath}
\usepackage{amsthm}
\usepackage{hyperref}

\usepackage[utf8x]{inputenc}
\usepackage{amssymb, amsmath, upgreek}
\newtheorem*{theorem*}{Theorem}

\newcommand{\Z}{\mathbb{Z}}
\newcommand{\Q}{\mathbb{Q}}
\newcommand{\R}{\mathbb{R}}
\DeclareMathOperator{\slope}{slope}
\DeclareMathOperator{\vol}{vol}
\DeclareMathOperator{\inj}{inj}

\begin{document}

\title{Mathematical reasoning and the computer.}
\author{Kevin Buzzard}
\email{k.buzzard@imperial.ac.uk}
\address{Department of Mathematics, Imperial College London}
\begin{abstract}
  Computers have already changed the way that humans do mathematics: they enable us to compute efficiently. But will they soon be helping us to \emph{reason}? And will they one day start reasoning themselves? We give an overview of recent developments in neural networks, computer theorem provers and large language models.
\end{abstract}

\maketitle

\section{Using computers to compute}

\subsection{Humans as computers}

In the early 1600s the Scottish mathematician John Napier constructed the first tables of logarithms. In his book (~\cite{napier}, p46) he reflected that ``the learned, who perchance may have plenty of pupils and computers'', would be able to take things further. You might think that having plenty of computers was rather uncommon in the 1600s, but back then the word ``computer'' referred to a human who would calculate.

The achievements of human computers should not be underestimated. Let us just give one example. Two such computers, Felkel and Vega, published tables of the factorizations of the positive integers up to $408000$ in 1776 and 1777\footnote{Although the title ``Tafel aller einfachen Factoren der durch 2, 3, 5 nicht theilbaren Zahlen von 1 bis 10 000 000'' of their work indicates that they had planned to go much further.}. As a consequence it became possible to create tables of the number of primes less than $N$, for $N$ in this range. Based on these tables, Legendre conjectured in the late 1700s that the number of primes less than $N$ was approximated by a function of the form $N/(C\log(N)+D)$. In the 1808 edition of his book ``Essai sur la th\'eorie des nombres''~\cite{legendre} on number theory he conjectured that $C=1$, and the statement of the prime number theorem was born (although the young Gauss had apparently independently come up with the same conjecture some years earlier). The theorem remained open for 100 years and its resolution, using complex analysis, by Hadamard and de la Vall\'ee Poussin was a triumph of late 19th century mathematics.

\subsection{Machines as computers}

Early electronic computers were the size of a house and orders of magnitude less powerful than a modern phone. By the time the University of Cambridge bought an EDSAC II machine in 1957 they were small enough to fit into a large room, and in those pre-transistor days they relied on bulky vacuum tubes (or ``valves'', as they were called in the UK) as switches. During the day, the EDSAC II was being used by oceanographers to do calculations which led the way to modern plate tectonic theory, but in the evenings the young Peter Swinnerton-Dyer would show up with a pile of punch cards and hand them to the operator; the machine would then spend all night counting the number of solutions $N_p$ to equations such as $y^2=x^3+x+37$ modulo prime numbers $p<100$ -- at least if they were lucky. Swinnerton-Dyer describes the days when the only output handed to him the next morning would be a printout saying ``error in punch card 4'', whereupon he would retire to his office to try and figure out where the offending hole was.

The result of this laborious process was a collection of data. At the time, data production was all the computer could do -- it was up to the humans to have the ideas. Birch and Swinnerton-Dyer tried to make rigorous sense of the idea that ``the more solutions a plane cubic had in the rationals, the more solutions it should have on average modulo prime numbers''. Let $A$ and $B$ be integers such that the complex roots of $x^3+Ax+B$ are distinct. It had long been known that the rational solutions to the plane cubic equation $y^2=x^3+Ax+B$, plus the ``point at infinity'', formed an abelian group. Mordell had proved that this group was of the form $\Z^r\times T$ with $T$ finite and $r$, the \emph{rank} of the cubic, a natural number. The insight of Birch and Swinnerton-Dyer, coming from the computations, was that if $p$ is prime and $N_p$ is the number of the solutions to the plane cubic mod~$p$, then the order of the growth of $\prod_{p\leq N}\frac{N_p}{p}$ should be $(\log N)^r$. This turned into the observation that the order of vanishing of the L-function of the curve at $s=1$ should be $r$, and the conjecture in its current form was born. It remains open.

The delineation of the tasks in the above stories is clear. The computer performs the computations, and then the mathematician takes over, providing their insights and interpretations. The computer is simply a tool with which one can collect evidence, at superhuman speed. Since the 1960s computers have become far smaller, far faster and far more available to the mathematician. Nowadays, using computers to collect experimental evidence in mathematics is commonplace. 

But generating more computational evidence for the Riemann Hypothesis or the Birch and Swinnerton-Dyer conjecture will not win you a Fields Medal. The currency of the pure mathematician is the \emph{theorem}. In this article we shall be concerned with new uses of computers, going beyond the merely mechanical work of computing for us. Can machines help us to think? It seems so! We will explain some examples. Can they even think for themselves? Yes -- in some sense -- but only at a very rudimentary level and only with certain kinds of problems. How far will things go? Nobody knows. Could machines one day start proving interesting deep conjectures by themselves? Some think so, others are more sceptical. Certainly there is no evidence of this happening right now -- currently this idea is science fiction. Rather than speculating about the future, we will restrict the scope of this article to a survey of what has happened, and what is happening currently.

There are three distinct topics which we shall discuss in the next three sections:

\begin{itemize}
\item Use of neural networks as a search tool for theorems, conjectures and counterexamples.
\item Automated and interactive theorem provers, and the mathematics they currently understand.
\item Large language models such as ChatGPT, and their efforts to do mathematics.
\end{itemize}

This article is written for mathematicians, and we will assume no background knowledge in any of these areas. In particular we will not discuss the technical advances in computer science which have enabled these tools to exist. Instead we will explain what these things \emph{do}, and how they are being used by mathematicians.

I strongly recommend Tom Hales' article ``Mathematics in the Age of the Turing Machine''~\cite{hales-turing}. One role of this article is to give an update of the area since Hales' article was written.

\section{Neural networks in mathematics}

In this section we discuss various uses of neural networks in mathematics. We start by giving a basic description of deep learning, one of the things a neural network can be used for.

\subsection{Introduction to deep learning}

The kind of problem which deep learning (or just ``learning'') attempts to solve is the following. We have two finite-dimensional real vector spaces $V=\R^m$ and $W=\R^n$, and a subset $S$ of $V$ which is the domain of a (non-linear) function $f:S\to W$. We have a finite table of pairs $(s,f(s))$ consisting of many elements $s\in S$ and their images under $f$. The table might have been expensive to calculate (it may have been computed using an algorithm which takes a long time to run, or it may have been produced manually by humans). What we would like to do is to make a computer program which can ``learn'' $f$, enabling us to compute its value on many more elements of $S$, quickly and with high accuracy. Let us give some examples.

A computer might internally represent a colour as an ``RGB value'', that is, as three numbers between 0 and 255 representing the amount of red, green and blue present in the colour. A $1000\times 1000$ digital photo can be represented as a list of million colours, and hence as a list of three million integers between 0 and 255. Thinking of these integers as real numbers, we can think of a digital photo as an element $s\in\R^{3000000}$. Let $S$ be the set of $1000\times 1000$ digital photos on the internet. We define $f:S\to\R$ by $f(s)=1$ if $s$ is a photo which contains a bicycle, and $f(s)=0$ otherwise. When a website wants to check that you are a human, it might give you a collection of photos and ask you which ones contain bicycles. The website might give many humans the same question, and define $f$ using the most popular answer. As time goes on, we generate a table of pairs. The question is whether we can now write a computer program which can accurately predict whether a given picture contains a bicycle.

Here is an example which might be more relevant to readers of this article. Mathematicians generate tables of data associated to mathematical objects. For example the L-functions and modular forms database~\cite{lmfdb} contains a very large list of elliptic curves over the rationals. An elliptic curve over the rationals has some kind of ``canonical'' minimal model, represented by 5 integers $(a_1,a_2,a_3,a_4,a_6)$ (the subscripts are to do with weights of the corresponding modular forms and the absence of $a_5$ is not a typo). To an elliptic curve over the rationals one can associate the rank (a natural number). So here we can set $V=\R^5$, we can let $S$ be the subset of $V$ corresponding to 5-tuples of integers such that the corresponding curve $Y^2+a_1XY+a_3Y=X^3+a_2X^2+a_4X+a_6$ is in ``canonical'' minimal form, and we can define $f(a_1,a_2,a_3,a_4,a_6)\in W=\R$ to be the rank of the corresponding elliptic curve. Our table of values of $f$ will be the table extracted from the L-functions and modular forms database. In this case we already have a computer program which can compute $f$, however the problem is that for some inputs it might be very slow, or may not terminate at all. The question is whether we can write a computer program which computes $f$ more quickly, with high accuracy.

Given $V$ and $W$ and the data of finitely many input and output values of $f$, an appropriate neural network will, after a period of ``training'', produce a function from $V$ to $W$ which is some kind of attempt to approximate or ``learn'' $f$ from the values we have trained on. The function is ``explicit'' in the sense that one can feed in numerical elements of $V$ and quickly get out elements of $W$. We will not go into the architecture of neural networks or an explanation of how this function is constructed.

To get an idea of how accurate the network's guess for $f$ is, the system is typically only trained on most of the available data (say, 90 percent). When the network has come up with a function interpolating the data it has seen, we can feed in the remaining 10 percent and compare the function's output with known results. A high success rate indicates that the network is accurately learning the function, and one can go on to now try running the function at points of $S$ not in our original dataset.

Geordie Williamson has been trying to understand the kind of mathematical problems which one might expect this sort of method to succeed. In some sense it is surprising that it succeeds at all, however over the last few years computers have become extremely good at deciding whether or not a given digital image contains a bicycle, so there is certainly some merit in the idea. A team led by Yang Hui He tried to use the techniques to compute ranks of elliptic curves~\cite{heellcurves1}, and was less successful. Williamson has observed that one key difference between the two problems is that the bicycle problem has a continuity property which the elliptic curve problem lacks: the output is unchanged by small changes in the input. If one slightly changes the colour of one pixel in a digital image, then the image remains essentially the same, so the answer to the question of whether it contains a bicycle will not change. However changing one coefficient of an elliptic curve can completely change the arithmetic of the curve; if you completely understand all the rational solutions to $y^2=x^3+37$ (a rank two curve) then this tells you nothing about the rational solutions to $y^2=x^3+38$ (a rank one curve). However Yang's team achieved a success rate of well over 90\% at rank prediction when instead of taking the Weierstrass equation as input, they took the first few hundred coefficients of the associated modular form. Yang's team has also had far more success (section 4.3.1 of~\cite{hecalabiyau}) in predicting whether the natural number invariant $h^{2,1}$ of a Calabi--Yau threefold over $\Q$ is greater than 50 or not; perhaps this is because the latter invariant is locally constant on the moduli space.

A tool which quickly predicts whether a certain invariant associated to a Calabi-Yau manifold is greater than 50, with 93 percent accuracy, can certainly be regarded as progress, but one can certainly argue that such a system is not producing \emph{theorems}, just different ways of doing computations. Let's now talk about three examples where neural networks have actually helped humans to \emph{prove new theorems}. One particularly nice fact about the work below is that in every case, mathematicians have been involved in the writing of the papers, and have ensured that the work is readable by mathematicians who may not have detailed knowledge of the theory of neural networks.

\subsection{Neural networks and knot theory}

The first example I'll discuss is the paper ``The signature and cusp geometry of hyperbolic knots'', by Davies, Juh\'asz, Lackenby and Tomasev~\cite{lackenbyknot}. This was a collaboration between two Oxford mathematicians and two computer scientists from the tech company DeepMind. The mathematicians are low-dimensional topologists, and the goal was to see if computers could help in the discovery of a new theorem in knot theory. There are countably infinitely many knots (up to ambient isotopy), and they can be tabulated in a reasonable way. Mathematicians over the years have discovered invariants of knots; some are ``discrete'' (for example the signature of a knot is an integer), and some are ``continuous'' (for example the volume of a (hyperbolic) knot is a real number). In fact there is a more fine-grained classification here: there are quantum invariants, topological invariants and gauge-theoretic invariants. There might be many relationships between these invariants, and the first question to ask is what kind of a theorem in this area would be interesting to mathematicians. One example of an interesting result might be some kind of link between continuous and discrete invariants, as this might be regarded as more profound.

With this in mind, the team used open source software to generate a huge table of knot invariants, and then used a neural network to try and spot hitherto unknown relationships between these invariants, and in particular to search for relationships between invariants with different origins. The team found that an appropriate neural network was able to accurately predict the signature of a knot (a discrete, cohomological invariant) given the data of the continuous invariants in the tables. The computer scientists on the team were then able to analyse the parameters of the neural network and isolate which of the continuous invariants were playing the most important role in the prediction. It was not possible to meaningfully understand ``in human terms'' the function which the neural network had learnt, however the knowledge of which continuous invariants were playing a major role was enough of a clue to continue. The baton was then passed on to the mathematicians, who came up with conjectural ideas relating the signature to the relevant continuous invariants. Their first conjecture was false, but they managed to prove the second one:

\begin{theorem*}[Davies, Juh\'asz, Lackenby, Tomasev] There exists a constant $c$ such that for any hyperbolic knot~$K$, $$|2\sigma(K)-\slope(K)|\leq c\vol(K)\inj(K)^{-3}.$$
\end{theorem*}

The theorem relates the signature $\sigma(K)$ of a knot to its slope, volume and injectivity radius. For more information about this result, I would recommend Lackenby's talk at the 2023 IPAM conference at UCLA~\cite{lackenbyIPAM}. Also highly recommended is Ernie Davis' rebuttal~\cite{davis-rebuttal} of the work, who argues that the role of deep learning in this story has been over-stated and that a conventional statistical analysis would probably have sufficed. Davis also has something to say about our next example, coming from representation theory.

\subsection{Neural networks and representation theory}

The same DeepMind team collaborated with other mathematicians on the combinatorial invariance conjecture, a problem in representation theory. Here we briefly discuss the results, contained in the paper~\cite{williamsonetal} by Blundell, Buesing, Davies, Velickovic and Williamson. Given a pair of elements of a finite symmetric group $S_n$, one can associate two invariants to the pair. The first, the Bruhat graph, is a directed graph which is typically a very unwieldy object, but is easy to compute. The second, the Kazhdan--Lusztig polynomial, is a very simple object, but is typically very hard to compute. The idea behind the conjecture is that the Kazhdan--Lusztig polynomial can (somehow) be computed from the Bruhat interval graph.

The basic idea behind the new approach is similar to the knot theory work. First one computes a large database of Bruhat interval graphs and Kazhdan--Lusztig polynomials. Next one encodes the graphs as sequences of numbers and thus as elements of a large real vector space. Next one tries to train a neural network to predict the Kazhdan-Lusztig polynomial. Once one has made the right design decisions, it turns out that the neural network can become good at this task. One then attempts to analyse the neural network for hints about what it is doing. Humans managed to distill from the network the idea that so-called extremal reflections in the graph were playing an important role in the prediction, which led them to a new definition in the theory, namely the concept of a hypercube decomposition (see section~3.4 of~\cite{williamsonetal}). This then led to a new formula for Kazhdan--Lusztig polynomials, which unfortunately depended on slightly more than the Bruhat decomposition graph so did not resolve the conjecture. However it also led to a new conjecture, which implies the combinatorial invariance conjecture and which the authors were able to check in over a million cases. 

Williamson says the following about this work: ``For me these findings remind us that intelligence is not a single variable, like an IQ number. Intelligence is best thought of as a multi-dimensional space with multiple axes: academic intelligence, emotional intelligence, social intelligence. My hope is that AI can provide another axis of intelligence for us to work with, and that this new axis will deepen our understanding of the mathematical world.''

Davis on the other hand in~\cite{davis-rebuttal} argues that although deep learning did play a key role in this work, it should be viewed as just ``another analytic tool in the toolbox of experimental mathematics rather than a fundamentally new approach to mathematics''. Although, as he goes on to say, ``How powerful a tool it is and how broadly applicable remains to be seen''. Time will tell.

\subsection{Searching for counterexamples in graph theory with neural networks}

Moving away from deep learning, another technique in this area is \emph{reinforcement learning}. In contrast to deep learning, where one has the data and wants to learn the function, in reinforcement learning one has the function (a ``reward function'') and one wants to construct appropriate data to maximise the reward. These sorts of techniques can be used to teach computers how to play video games, for example. But they can also be used to search for interesting mathematical objects. Adam Wagner used reinforcement learning to search for counterexamples in graph theory. We cite one example from his beautiful paper~\cite{wagner}. A conjecture from 2010 stated that if $G$ is a connected graph with $n\geq3$ vertices, having largest eigenvalue $\lambda_1$ and matching number $\mu$, then $\lambda_1+\mu\geq 1+\sqrt{n-1}$. Wagner considered graphs on $n=19$ vertices, with the source vector space $V$ encoding the presence of edges between these vertices. The network continually modified the input graph, attempting to minimise $\lambda_1+\mu$. Eventually the network found a graph for which $\lambda_1+\mu$ was so small that the inequality was violated. There are several other examples in the paper, including illustrative examples where the network did not directly find a counterexample to a given question, but started to make graphs which had a clear structure, enabling the human to take over and finish the job. An overview of the work is given in Wagner's IPAM talk~\cite{wagnerIPAM}.

The above examples indicate novel applications of these new tools, however it might be the case that applicability of these tools is limited to areas where computation is possible and tables exist, so they might not be useful if you are interested in, say, perfectoid spaces. In the next section however, we discuss tools which seem to be able to access all of pure mathematics.

\section{Automated and interactive theorem provers}

Our second example of new uses of computers in mathematics is an overview of the capabilities of ATPs (automated theorem provers) and ITPs (interactive theorem provers). These are both systems which know or can be told various axioms of a theory, and have a language rich enough to express theorems in the theory. The job of an automated theorem prover is to attempt to automatically generate proofs of theorems in the theory. The job of an interactive theorem prover is to offer a powerful front end where humans can type in their own proofs of theorems and the system will check that they are valid. 

Despite the superficial similarities between ATPs and ITPs, in practice they are used to perform very different tasks. Let us first give a brief overview of ATPs.

\subsection{Automated theorem provers.}

Automated theorem provers are typically used to prove (possibly very complex) theorems in a simple theory, perhaps using first order logic. An example might be the first order theory of groups. The system is given the axioms for a group, and a formula which is true in all groups, for example $(ab)^{-1}=b^{-1}a^{-1}$, and it attempts to prove the theorem from the axioms. These systems play an important role in software verification, but we shall focus on their applications in mathematics.

One of the early successes of these systems was a proof of the Robbins conjecture. In the 1930s, Robbins came up with a collection of axioms which he conjectured were an axiomatisation of Boolean algebras; the hard part was verifying the axioms of a Boolean algebra from Robbins' axioms. William McCune proved Robbins' conjecture in 1996, using an ATP; the story made the New York Times. Allen Mann's write-up~\cite{mannrobbins} contains a human-readable 14 page account of McCune's computer proof: in particular the output of the ATP is comprehensible to a human in finite time. Within a year, Bernd Dahn had simplified the proof and the resulting seven page paper~\cite{dahnrobbins} was published in the Journal of Algebra.

However these systems are no longer limited to discovering short proofs. In 2012 Phillips and Stanovsk\'y, and Waldmeister (an ATP) proved~\cite{bruckloops} that a Bruck loop\footnote{A loop is basically a group with the associativity axiom dropped.} with abelian inner mapping group has nilpotency class at most 2. The proof generated by the ATP was over 30,000 lines long and when written out in a locally human-readable format (surely no human would ever read the whole thing) fills just over 1000 pdf pages. Note of course that the proof was not found by ``brute force search'' -- these systems are far cleverer. We cannot however rule out the existence of a far shorter, and perhaps even human-comprehensible, proof.

Whilst the proofs which these systems can discover can be huge, the \emph{nature} of these proofs seems to be inherently very ``low-level''. A common usage nowadays for automated theorem provers is as components of interactive theorem provers, which we now describe.

\subsection{Interactive theorem provers.}

Like automated theorem provers, interactive theorem provers (ITPs) know the rules of a logic and the axioms of a theory, and can be used to prove theorems in the theory. However typically now the logic and theory are richer and more complex, making it practical to formalise cutting-edge modern mathematics. One can think of an ITP as a programming language where the code corresponds to mathematical definitions, theorems and proofs. The big difference between interactive and automated theorem provers is that in an ITP, the user is expected to type in the key ideas of the proof themself. A modern ITP will typically have tactics, which are little computer programs capable of filling in steps which are clear to a mathematician but which would be tedious to prove from the axioms directly. For example a proof that $(x+y)^3=x^3+3x^2y+3xy^2+y^3$ directly from the axioms of a ring is surprisingly long, because for example the step after expanding out the brackets -- ``rearrange these eight terms into the right order and put all the brackets in the right place'' -- corresponds to many many applications of commutativity and associativity of addition and multiplication. A ring theory tactic would perform this computation in just one line and hide these tedious arguments from the user. Tactics make the process of teaching more advanced proofs to the ITP feasible in a reasonable time. However it is currently still far more time-consuming than writing the analogous results in a LaTeX document, even for expert users. Bringing down the time it takes to write a proof in an ITP to something nearer to the time it takes to write it on paper would surely increase take-up of these systems by mathematicians. Novel ideas are being tried all the time. Paulson and Blanchette created a ``hammer'' tactic for the Isabelle/HOL ITP, which attempts to prove intermediate results in the ITP by sending them off to an ATP and then attempting to construct a proof valid in the ITP from the output. Abby Goldberg, Druv Bhatia and Rob Lewis wrote the {\tt polyrith} tactic for Lean which calls an instance of the Sage computer algebra system on the cloud to answer a certain question in algebra, and then gets Lean to formally check the answer. Finally Scott Morrison has written a general purpose {\tt sagredo} tactic for Lean which establishes a dialogue with a large language model (more on these later) and attempts to use it to generate a Lean proof.

Historically, ITPs were used, by computer scientists, to verify basic results in undergraduate level mathematics. This century users became more ambitious. In 2004 Jeremy Avigad led a team which verified the prime number theorem in Isabelle, arguably the first ``serious'' mathematical result verified in an ITP. Within a few years there had been several more. Georges Gonthier showed that the Coq theorem prover was powerful enough to formally verify the four colour theorem. The original Appel--Haken proof was a computer-assisted proof, in the sense that it relies at some point on a brute force computation which is too large to be done by humans. Indeed every currently known proof of the four colour theorem is computer-assisted. This raises the question of whether the computer program used to finish the job has bugs. Gonthier's work is a full proof of the result in the Coq theorem prover, and in particular the computer-assisted part of the calculation is verified as bug-free, according to the prover. This work was a tour de force at the time. One could be paranoid and suggest that we have reduced the proof of the 4 colour theorem to the assertion that Coq has no bugs. However we have done more than this: one can run the compiled output of Gonthier's code through an independent type checker, a very simple program which one can check by hand and which just answers the question ``does this output correspond to a proof of this theorem from these axioms?'' (in contrast to an ITP, which contains a wide array of tools for actually constructing proofs). If the type checker is happy with the proof produced by Coq then the question of whether Coq's kernel has bugs is now no longer relevant. And if you are concerned about, for example, the compiler used to compile the type checker, or the chipset of the computer used to run it, then you can just write another typechecker in a different language and run it on a different computer. At any rate, you can choose your level of paranoia; the conclusion of the work is at the very least the assertion that the proof of the 4 colour theorem has now been checked to a \emph{far} higher standard than before Gonthier's work.

Having demonstrated that ITPs could be used to check computer-assisted proofs, a team led by Gonthier went on to demonstrate that they were also capable of checking 20th century Fields Medal level mathematics. The team typed a full proof of the Feit--Thompson odd order theorem into Coq, over a period of six years. The original proof of the theorem was part of the reason Thompson was awarded the Fields Medal in 1970.

Thomas Hales then showed that the systems were powerful enough to handle a modern computer-assisted proof. His proof with Ferguson~\cite{haleskepler} of the Kepler conjecture was computer-assisted, and the Annals would only publish the result with a disclaimer saying that the referees were not able to say with 100 percent confidence that the proof was complete. Hales' response was to put together a team of people who formalised the entire proof in an ITP and thus put any question of a bug beyond all reasonable doubt. The disclaimer has since been removed.

More recently, the Lean community have been demonstrating that it is now possible to formalise modern research level mathematics ``in real time''. Recent examples are the 2019 Dahmen--H\"olzl--Lewis formalisation~\cite{dhl} of the 2016 Ellenberg--Gijswijt solution to the cap set problem~\cite{ellenberg-gijswijt}, the 2022 Gadgil--Tadipatri formalisation~\cite{gadgilweb} of Gardam's 2021 disproof~\cite{gardam} of the Kaplansky unit conjecture, and the 2022 Bloom--Mehta~\cite{unitfractionsGH} formalisation of Bloom's 2021 resolution~\cite{bloom} of a conjecture of Erd\H{o}s and Graham on continued fractions. Note that in 2021 Thomas Bloom knew nothing about Lean, but he was introduced to it by Bhavik Mehta and between them they formalised the full proof before the paper had even got a referee's report. There is currently ongoing work of Mehta and Ya\"el Dillies (currently an undergraduate in Cambridge) formalising important 2022 and 2023 results in additive combinatorics, adding to the evidence that, at least in some areas of mathematics, real-time formalisation is becoming a reality.

The results cited above are certainly complex mathematics, but the arguments remain entirely in the domain of ``low-level'' (prime numbers, planar graphs, finite groups, balls in 3-space, fractions,\ldots). More recently the Lean community has been attempting to engage with modern, far more subtle, mathematical objects such as perfectoid spaces and the homological algebra of condensed abelian groups. Johan Commelin and Adam Topaz successfully led a project called the ``Liquid Tensor Experiment''~\cite{scholze-expmath} whose goal was to formalise a proof of a theorem of Clausen and Scholze in the Lean ITP; see~\cite{commelinblog} and~\cite{commelin-topaz} for more details. One notable consequence of the work was that the dependency on the theory of stable homotopy groups of spheres was removed during the formalization process; the computer kept track of precisely what was needed from this area, and ultimately it turned out that one could make do with a lot less than the theorem of Breen and Deligne initially cited by Clausen and Scholze.

Those concerned about whether the nature of these systems mean that they are limited to algebraic rather than geometric results should rest assured: Massot, van Doorn and Nash have formalised a proof in Lean of sphere eversion~\cite{sphereeversion}, the fact that one can turn a sphere inside-out in 3-space (allowing it to pass through itself, but with creasing or cutting not allowed). The success of these projects is an indication that the sky now seems to be the limit when it comes to the mathematics which these systems can handle, although there is still plenty of scope for formalisation of geometric results; in particular arguments which heavily rely on pictures will present an interesting challenge. A related question is how long the process of formalisation takes, and whether this time varies between subfields (in particular, whether formalising geometry takes longer than formalising arithmetic). Another important question is how to decrease this time.

The Liquid Tensor Experiment and the sphere eversion project were both built on Lean's mathematics library {\tt mathlib}; this is a library which now contains essentially all of a standard undergraduate mathematics degree, as well as many results at Masters level in algebra, number theory and algebraic geometry. The library grows daily, and more and more mathematicians are getting involved. One way of thinking about the library is that it is a 21st century Bourbaki, always working in a large generality and focusing mostly on theorems rather than examples.

ITPs and ATPs turn mathematics into a game, like chess or go. Breakthroughs in AI have created machines which are superhuman at both chess and go, and this raises the possibility that future AIs will become superhuman at mathematics, proving theorems which humans are interested in but which they cannot solve themselves. This idea is still science fiction right now; whilst AIs can write code in these languages, their current level is that of a strong schoolchild or average 1st year undergraduate. We highlight two of the difficulties present in translating the successes of the game domain into mathematics. Firstly, mathematics is a single-player game rather than a two player game; it is difficult to ascribe a ``score'' to a partial proof of a theorem, other than the obvious ``0\%'' if you haven't finished and ``100\%'' if you have; pure mathematicians may well not be interested in a theorem which is ``95\% proved'', if this even has any meaning. And second, mathematics has an infinite and high-dimensional action space, by which I mean that at any point in a proof there are infinitely many things which you can do next, including applying any one of thousands of applicable lemmas, many with large numbers of parameters. It is right now hence difficult to ``guide'' the systems towards a successful proof. In the final section we will discuss large language models such as ChatGPT, and in particular ask whether these systems are capable of helping with this problem.

Here are some more realistic near-term goals for these systems. Firstly, one could imagine them powering interactive error-free textbooks. Recent work of Massot and Miller~\cite{massotIPAM} has shown that this is already becoming a reality; they have created a system which takes as input a Lean proof and outputs a web page containing a human natural language proof. The viewer can then interact with the proof, and ``unfold'' arguments right down to the axioms of the system. This is only one carefully-curated example but it is strong evidence that the technology is ready for a much larger project.

Secondly, one could imagine ``chatbots'' who are experts in a given domain such as algebraic geometry, backed up by a database such as a formalised version of the Stacks Project~\cite{stacks-project} and where students are able to query the system for theorems, examples and counterexamples. If the systems work in the language of the theorem prover in question, then the user would have to learn this programming language. We teach mathematics undergraduates python -- it is not unreasonable to imagine that we could also be teaching them a language such as Lean (and indeed such courses are starting to appear around the world). However, such a system would be much easier to use if the bots were able to understand queries and respond in English, so perhaps it is time to end this section and begin to talk about recent developments in AI generation of natural language.

\section{Large language models}

Large language models are neural network-based systems which provide a probability distribution as an answer to the question ``what is the next token in this sequence of tokens?'' (where a token might be a word or perhaps a letter or number). Applications of this technique include things such as ``write a paragraph of text answering a given question'' and ``write computer code in the language of an ITP proving a given theorem''. Recent breakthroughs in this area have got the systems to the state where they are capable of writing coherent, correct and relevant English sentences, something which a few years ago was a hard unsolved problem. Right now the most famous of these systems is undoubtedly OpenAI's ChatGPT~\cite{chatgpt}, a large language model which one can access for free on the internet and which seems to have opened the eyes of the general public to the current power of these systems. The systems are trained on extremely large bodies of text (e.g. the entire internet) and are now able to respond to many questions ``in the way a human would''. Computer scientists inform us that progress in this area is ``exponentially fast'' right now, and some would like to infer from this that progress will continue to be exponential for some time, which will presumably quickly render the contents of this section out of date. For now, let us ask: can these things currently do mathematics?

If you ask ChatGPT to prove that there are infinitely many primes, then (because it is trained on the internet) it will happily rattle off some variant of Euclid's proof. It may get a little confused in its treatment of the case where $1+\prod_{i=1}^np_i$ is composite, but one could imagine that a random proof that one finds on the internet might also be slightly confused at this point. However, what happens when we ask it for a proof of the much harder result that there are infinitely many primes which end in 7? Like a first year undergraduate faced with this question, it thinks about the proof which it knows of the infinitude of primes, and tries to generalise it to this situation. Unfortunately it is not true that an arbitrary number of the form $10\prod_i p_i+7$ must have a prime factor which ends in 7, but this does not stop the model from confidently arguing that this is the case, like the undergraduate might. Worse -- if asked to prove that there are infinitely many even numbers which end in 7, the system might well try a similar strategy and write a nonsensical paragraph. The reader should feel free to try these examples themselves in whatever large language model they have access to, to get a feeling for what these things can or cannot do right now.

At the time of writing, a big problem with these systems when it comes to writing mathematics in a ``natural language'' such as English is that they will happily assert false statements. Multiple systems now exist, and some (for example GPT4~\cite{gpt4}) have performed competently in difficult school level multiple choice mathematics exams such as AMC12. Here a false statement is not the end of the world -- it might lose you a mark, but you can still certainly end up passing the exam. But one false assertion in a computer-generated proof of the Riemann hypothesis and the entire edifice comes tumbling down so, if we are to take these systems further, then unjustified or incorrect claims need to be eliminated. Some people hope that better training will somehow drastically lessen the likelihood of the system emitting false statements, but we are not there yet.

A different approach would be to instead start training the systems to write not in natural languages such as English but in the language of an ITP. If a system generates code then this code can be run through the ITP and the system can be immediately informed if they are talking nonsense. Writing code which compiles and corresponds to a mathematical proof is much harder than writing an English language text which can be passed off as a proof, not least because proofs in an ITP must leave no stone unturned. Both Meta and OpenAI have produced work in this direction recently (\cite{polu2022formal}, \cite{facebookHTPS}). Both systems have managed to automatically generate Lean code corresponding to proofs of theorems at olympiad level (see the papers for more details of what has actually been achieved). One challenge, surely still at least a few years away, is the IMO Grand Challenge: to write a system which can get a gold medal in the International Mathematics Olympiad. Whilst questions which can be solved by very smart schoolchildren are still a long way from mathematical research, it would still represent a very impressive milestone if the systems could be pushed this far. It is not at all clear to the author whether success is more likely to come from natural language systems or ITP-backed systems -- or perhaps even from a hybrid approach. One issue with an ITP-based approach is that the questions will have to be translated into the language of the ITP, and it is not always clear how this should be done: for example a question of the form ``find the smallest natural number with property $X$'' is clearly expecting an answer of the form ``37'' rather than one of the form ``it's $n$, where $n$ is the smallest natural number with property $X$'', an answer which is logically correct but not what is intended. It is not immediately clear what to do here without leaking information which the human candidates will not have. Another issue is that training is difficult: there are solutions to all IMO problems on the internet already, so any system which has been trained on the internet has already seen all of the answers to all of the questions, and questions at the appropriate level which are not already publically available can be difficult to come by.

\section{Summary}

Machines have already changed mathematics, by helping us to compute more quickly; this happened decades ago. But now we seem to be at the dawn of a new era, where computers are able to engage with the concept of proof. Neural networks have helped human mathematicians to discover new theorems and new counterexamples. ITPs have helped humans to simplify the proofs of recent results in the literature, and in some areas of mathematics (for example additive combinatorics), formalizing a modern paper in a matter of months is now often possible. Large language models have currently had little effect on mathematics beyond school level, but \emph{if} the current rate of improvement in the area continues, then they too will one day be playing a role in mathematical research.

I will end with the following observation. A lot of the research in this area is coming out of the tech companies or computer science departments; indeed, in the recent past, few mathematics departments have hired in this or related areas, and so researchers are more likely to be found in computer science departments. Publications in this area are often found in journals unfamiliar to mathematicians. Things are beginning to change though: DeepMind has collaborated with mathematicians from Oxford and Sydney, and courses on how to use an ITP are being taught in mathematics departments at Paris Saclay, Exeter, Fordham, Imperial College London, UCL, Carnegie--Mellon and Harvard, and others are springing up every year. It is essential that mathematicians remain at the forefront of current developments however, so that the area is being guided by experts and remains relevant and representative of the mathematics which is currently happening in our departments.

\section{Acknowledgements}

The author would like to thank Patrick Massot, Michael Harris, Siddharta Gadgil, Filippo Nuccio, Oliver Nash, Andrew Granville and the anonymous referees for all their feedback. Any errors or omissions remaining are of course the fault of the author.

\bibliographystyle{amsalpha}

\newcommand{\etalchar}[1]{$^{#1}$}
\providecommand{\bysame}{\leavevmode\hbox to3em{\hrulefill}\thinspace}
\providecommand{\MR}{\relax\ifhmode\unskip\space\fi MR }
\providecommand{\MRhref}[2]{%
  \href{http://www.ams.org/mathscinet-getitem?mr=#1}{#2}
}
\providecommand{\href}[2]{#2}

\end{document}